\documentclass[11pt,a4paper]{article}
\usepackage[hyperref]{emnlp-ijcnlp-2019}
\usepackage{times}
\usepackage{latexsym}
\usepackage{amsmath,amssymb,amsfonts}
\usepackage{algorithmic}
\usepackage{graphicx}
\usepackage{url}
\usepackage{multirow}
\usepackage{booktabs}
\usepackage{bm}
\usepackage{enumitem}
\usepackage{subfigure}

\def\e{{\bf e}}

\def\x{{\bf x}}

\def\0{{\bf 0}}
\def\1{{\bf 1}}

\def\CM{{\mathcal C}}

\def\JM{{\mathcal J}}

\def\MM{{\mathcal M}}

\def\SM{{\mathcal S}}

\newcommand{\nop}[1]{}

\aclfinalcopy % Uncomment this line for the final submission

\usepackage{color}

\title{What You See is What You Get: \\Visual Pronoun Coreference Resolution in Dialogues}
\author{
Xintong Yu$^\heartsuit$\thanks{$\quad$Equal contribution.},~ Hongming Zhang$^\clubsuit$$^{*}$, Yangqiu Song$^\clubsuit$, Yan Song$^\spadesuit$, and Changshui Zhang$^\heartsuit$\\
$^\heartsuit$Institute for Artificial Intelligence, Tsinghua University (THUAI), \\Department of Automation, Tsinghua University\\
$^\clubsuit$Department of CSE, The Hong Kong University of Science and Technology\\
$^\spadesuit$Sinovation Ventures\\
yuxt16@mails.tsinghua.edu.cn, hzhangal@cse.ust.hk, yqsong@cse.ust.hk, \\clksong@gmail.com, zcs@mail.tsinghua.edu.cn
}

\date{}

\begin{document}
\maketitle
\begin{abstract}
Grounding a pronoun to a visual object it refers to requires complex reasoning from various information sources, especially in conversational scenarios.
For example, when people in a conversation talk about something all speakers can see, they often directly use pronouns (e.g., it) to refer to it without previous introduction.
This fact brings a huge challenge for modern natural language understanding systems, particularly conventional context-based pronoun coreference models.
To tackle this challenge, in this paper, we formally define the task of visual-aware pronoun coreference resolution (PCR) and introduce VisPro, a large-scale dialogue PCR dataset, to investigate whether and how the visual information can help resolve pronouns in dialogues.
We then propose a novel visual-aware PCR model, VisCoref, for this task and conduct comprehensive experiments and case studies on our dataset.
Results demonstrate the importance of the visual information in this PCR case and show the effectiveness of the proposed model.
\end{abstract}

\section{Introduction}
% 1.5 pages

The question of how human beings resolve pronouns has long been an attractive research topic in both linguistics and natural language processing (NLP) communities, for the reason that pronoun itself has weak semantic meaning~\cite{ehrlich-1981-search} and the correct resolution of pronouns requires complex reasoning over various information.
As a core task of natural language understanding, pronoun coreference resolution (PCR) \cite{hobbs1978resolving} is the task of identifying the noun (phrase) that pronouns refer to. 
Compared with the general coreference resolution task, the string-matching methods are no longer effective for pronouns \cite{stoyanov-etal-2009-conundrums}, which makes PCR more challenging than the general coreference resolution task.

\begin{figure}[t]
  \centering
  \includegraphics[width=0.5\textwidth]{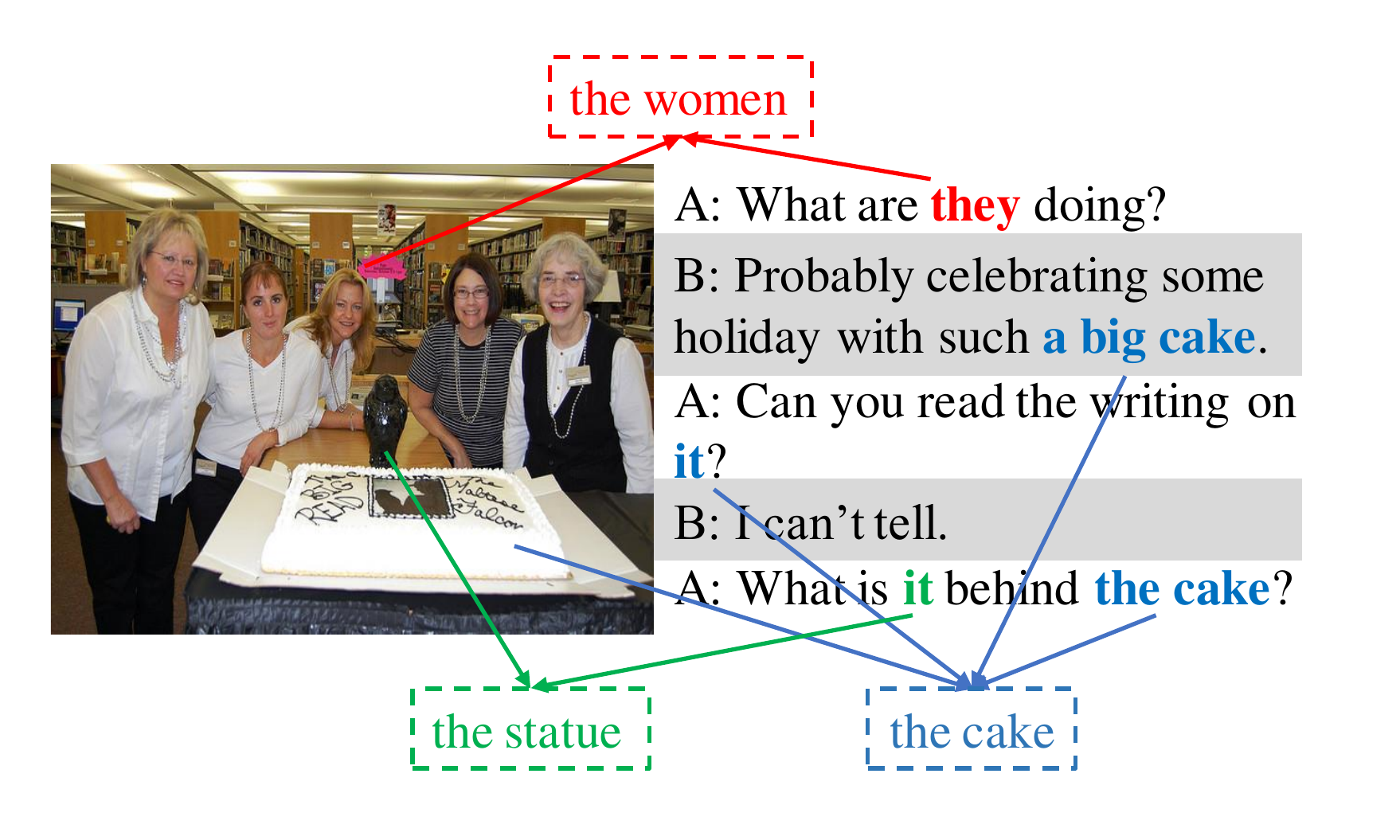}
  \caption{An example of a visual-related dialogue. Two people are discussing the view they can both see. Pronouns and noun phrases referring to the same entity are marked in same color. The first ``it'' in the dialogue labeled with blue color refers to the object ``the big cake'' and the second ``it'' labeled with green color refers to the statue in the image. }
  \label{fig:visdial}
%   \vskip -1em
\end{figure}

Recently, great efforts have been devoted into the coreference resolution task~\cite{raghunathan2010multi,clark2015entity,clark-manning-2016-deep,lee-etal-2018-higher} and good performance has been achieved on formal written text such as newspapers \cite{pradhan2012conll,DBLP:conf/naacl/ZhangSS19} and diagnose records~\cite{DBLP:conf/acl/ZhangSSY19}.
However, when it comes to dialogues, where more abundant information is needed, the performance of existing models becomes less satisfying.
The reason behind is that, different from formal written language, correct understanding of spoken language often requires the support of other information sources.
For example, when people chat with each other, if they intend to refer to some object that all speakers can see, they may directly use pronouns such as ``it'' instead of describing or mentioning it in the first place.
Sometimes, the object (name or text description) that pronouns refer to may not even appear in a conversation, and thus one needs to ground the pronouns into something outside the text, which is extremely challenging for conventional approaches purely based on human-designed rules~\cite{raghunathan2010multi} or contextual features~\cite{lee-etal-2018-higher}.

% One paragraph of case study
A visual-related dialogue is shown in Figure~\ref{fig:visdial}.
Both A and B are talking about a picture, in which several people are celebrating something.
In the dialogue, the first ``it'' refers to the ``the big cake,'' which is relatively easy for conventional models, because the correct mention just appears before the targeting pronoun.
However, the second ``it'' refers to the statue in the image, which does not appear in the dialogue at all.
Without the support of the visual information, it is almost impossible to identify the coreference relation between ``it'' and ``the statue.''

In this work, we focus on investigating how visual information can help better resolve pronouns in dialogues. 
To achieve this goal, we first create VisPro, a large-scale visual-supported PCR dataset.
Different from existing datasets such as ACE~\cite{nist2003ace} and CoNLL-shared task~\cite{pradhan2012conll}, VisPro is annotated based on dialogues discussing given images.
In total, VisPro contains annotations for 29,722 pronouns extracted from 5,000 dialogues. 
Once the dataset is created, we formally define a new task, visual pronoun coreference resolution (Visual PCR), and design a novel visual-aware PCR model VisCoref, which can effectively extract information from images and leverage them to help better resolve pronouns in dialogues.
Particularly, we align mentions in the dialogue with objects in the image and then jointly use the contextual and visual information for the final prediction.
% use extracted local contextual information as attention to acquire useful information from the image

The contribution of this paper is three-folded: (1) we formally define the task of visual PCR; (2) we present VisPro, the first dataset that focuses on PCR in visual-supported dialogues; (3) we propose VisCoref, a visual-aware PCR model.
Comprehensive experiments and case studies are conducted to demonstrate the quality of VisPro and the effectiveness of VisCoref.
The dataset, code, and models are available at:
\url{https://github.com/HKUST-KnowComp/Visual\_PCR}.
% https://github.com/HKUST-KnowComp/Visual\_PCR.

\section{The VisPro Dataset}\label{sec:dataset}

\begin{figure}
    \centering
    \includegraphics[width=\linewidth]{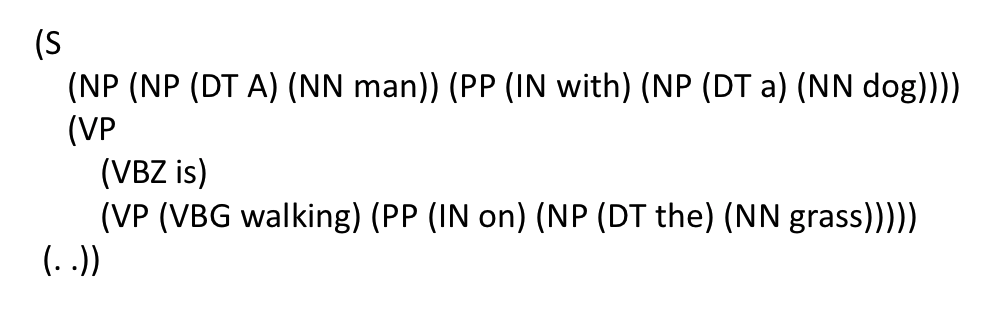}
    \caption{Example syntax parsing result of the sentence ``A man with a dog is walking on the grass.''}
    \label{fig:syntex_tree}
\end{figure}

To generate a high-quality and large-scale visual-aware PCR dataset, we select VisDial~\cite{visdial} as the base dataset and invite annotators to annotate.
In VisDial, each image is accompanied by a dialogue record discussing that image. One example is shown in Figure~\ref{fig:visdial}.
In addition, VisDial also provides a caption for each image, which brings more information for us to create VisPro\footnote{The information contained in the caption only helps provide noun phrase candidates for workers to annotate and will not be treated as part of the dialogue.}.
In this section, we introduce the details about the dataset creation in terms of pre-processing, survey design, annotation, and post-processing.

\subsection{Pre-processing}
To make the annotation task clear to annotators and help them provide accurate annotation, we first extract all the noun phrases and pronouns with Stanford Parser \cite{Manning2003Accurate} and then provide the extracted noun phrases as candidate mentions to annotate on.
To avoid the overlap of candidate noun phrases, we choose noun phrases with a height of two in parse trees. One example is shown in Figure~\ref{fig:syntex_tree}. In the syntactic tree for the sentence ``A man with a dog is walking on the grass,'' we choose ``A man,'' ``a dog'' and ``the grass'' as candidates. If the height of noun phrases is not limited, then the noun phrase ``A man with a dog'' will cover ``A man'' and ``a dog,'' leading to confusion in the options.

\begin{figure}[t]
    \centering
    \includegraphics[width=\linewidth]{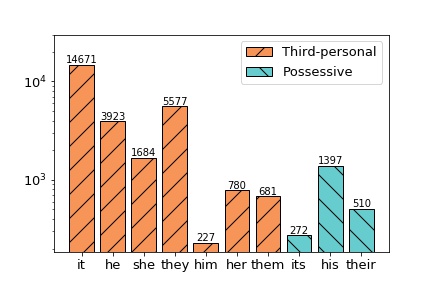}
    \caption{Distribution of pronouns in VisPro.}
    \label{fig:stat}
\end{figure}

Following \cite{strube2003machine,ng2005supervised}, we only select third-person personal (it, he, she, they, him, her, them) and possessive pronouns (its, his, her, their) as the targeting pronouns.
In total, the VisPro dataset contains 29,722 pronouns of 5,000 dialogues selected from 133,351 dialogues in VisDial v1.0~\cite{visdial}.
We choose dialogues in which the number of pronouns ranges from four to ten for the following reasons. For one thing, dialogues with few pronouns are of little use to the task. For another, dialogues with too many pronouns often contain repeating pronouns referring to the same object, which makes the task too easy. The dialogues selected contain 5.94 pronouns on average.
Figure~\ref{fig:stat} presents the distribution of different pronouns. From the figure we can see that ``it'' and ``they'' are used most frequently in the dialogues. 

\subsection{Survey Design}

We divide 29,722 pronouns from 5,000 dialogues into 3,304 surveys. 
In each survey, besides the normal questions, we also include one checkpoint question to control the annotation quality\footnote{We design the checkpoint dialogue straightforward and unambiguous such that any responsive worker can easily provide the correct annotation.}. 
In total, each survey contains ten questions (nine normal questions and one checkpoint question). Each question is about one pronoun.

The survey consists of three parts. We begin by explaining the task to the annotators, including how to deal with particular cases such as multi-word expressions. Then, we present examples to help the annotators better understand the task. Finally, we invite them to provide annotations for the pronouns in the dialogues.

\begin{figure}[t]
  \centering
  \includegraphics[width=\linewidth]{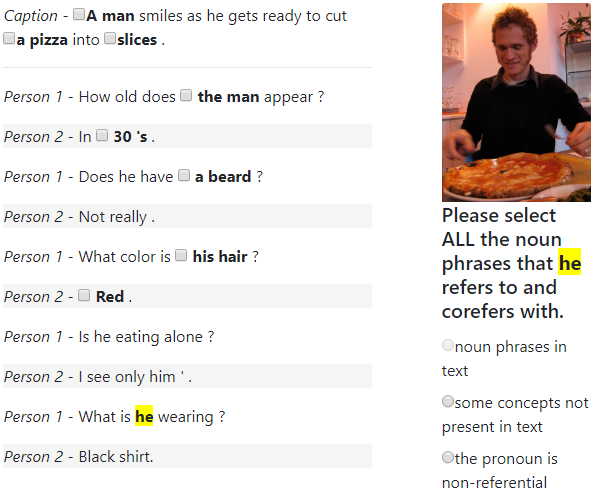}
  \caption{Annotation interface on MTurk. Workers are asked to select the anaphoricity type of the highlighted pronoun in the right panel and all the antecedents of an anaphoric pronoun on the left panel.}
  \label{fig:interface}
\end{figure}

One example of the annotation interface is shown in Figure~\ref{fig:interface}.
The text and the image of the dialogue are displayed on the left and right panel, respectively. 
For each of the targeting pronoun, the workers are asked to select all the mentions that it refers to.
% Each question asks the worker to select all the noun phrases that a highlighted pronoun refers to and corefers with from several predefined options on the left panel. 
If any of the noun phrases is selected, the reference type of the pronoun on the right panel will be set to ``noun phrases in text'' automatically, and vice versa. If the concept that the pronoun refers to is not available in the options, or the pronoun is not referring to anything in particular, workers are asked to choose ``some concepts not present in text'' or ``the pronoun is non-referential'' on the right panel accordingly. The nine normal questions in each survey are consecutive so that pronouns in the same dialogue are displayed sequentially.

\subsection{Annotation}
% Introduce the selection of participants and how do annotate it (e.g., for each survey, how many people do we invite for each survey)

We employ the Amazon Mechanical Turk platform (MTurk) for our annotations.

We require that our annotators have more than 1,000 approved tasks and a task approval rate more than 90\%. They are also asked to pass a simple test of pronoun resolution to prove that they understand the instruction of the task.
Based on these criteria, we identify 186 valid annotators. 
For each dialogue, we invite at least four different workers to annotate.
In total, we collect 122,443 annotations at a total cost of USD 3,270.80. 
We support the multiple participation of annotators by ensuring that subsequent surveys are generated with their previously-unlabelled dialogues.

\subsection{Post-processing}\label{sec:post-processing}
% Introduce how do you post process the annotated data to generate the final dataset. 

Before processing the annotation result, we first exclude the annotation of workers who fail to answer the checkpoint questions correctly.
As a result, 116,300 annotations are kept, which is 95\% of the overall annotation results.
We then decide the gold mentions that pronouns refer to using the following procedure:

\begin{itemize}[leftmargin=*]
    \item Step one: We look into the annotations of each worker to find out the coreference clusters he annotates for each dialogue. To achieve this goal, we merge the intersecting sets of noun phrase antecedents for pronouns in the same dialogue. We observe that annotators often make the right decision for noun phrases near the anaphor pronoun, but neglect antecedents far away. It also happens in the annotation of other coreference datasets \cite{Chen2018PreCo}. Therefore, we generate clusters from different pronouns in the same dialogue rather than merging antecedents for each pronoun separately.
If an entity is mentioned and referred to by pronouns for multiple times in the dialogue, combining the antecedents for all pronouns could create a more accurate coreference cluster for the entity.
    \item Step two: We adjudicate the coreference clusters for the dialogue by majority voting within all workers.
    \item Step three: We then decide the anaphoric type of all pronouns by voting. If a pronoun is considered to refer to somef noun phrases in the text, we find out the coreference cluster it belongs to and choose the noun phrases in the cluster that precede it as its antecedents.
    \item Step four: We randomly split the collected data into train/val/test sets of 4,000/500/500 dialogues, respectively.
\end{itemize}

After collecting the data, we found out that 73.43\% of pronouns act as an anaphor to some noun phrases, 5.67\% of pronouns do not have a suitable antecedent, and the rest 20.90\% are not referential.
Among all the pronouns that have noun phrases as antecedents, 13.45\% of them do not have an antecedent in the dialogue context.\footnote{The antecedent labeled by the worker is provided by the caption.}
For anaphoric pronouns, each has 2.06 antecedents on average.
In the end, we calculate the inner-annotator agreement (IAA) to demonstrate the quality of the resulting dataset.
Following conventional approaches \cite{pradhan2012conll}, we use the average MUC score between individual workers and the adjudication result as the evaluation metric.
The final IAA score is 72.4, which indicates that the workers can clearly understand our task and provide reliable annotation.

\section{The Task}\label{sec:task}
In this work, we focus on jointly leveraging the contextual and visual information to resolve pronouns.
Thus, we formally define the visual-aware pronoun coreference resolution as follows:

Given an image $I$, a dialogue record $D$ which discusses the content in $I$, and an external mention set $\MM$,
for any pronoun $p$ that appears in $D$, the goal is to optimize the following objective:
\begin{equation}\label{Objective}
\JM = \frac{\sum_{c \in \CM}{e^{F(c, p, I, D)}}}{\sum_{s \in \SM}e^{F(s, p, I, D)}},
\end{equation}
where $F(\cdot)$ is the overall scoring function of $p$ refers to $c$ given $I$ and $D$. $c$ and $s$ denote the correct mention and the candidate mention, and $\CM$ and $\SM$ denote the correct mention set and the candidate mention set, respectively.
Note that in the case where no golden mentions are annotated, the union of all possible spans in $D$ and $\MM$ are used to form $\SM$.

\begin{figure}[t]
  \centering
  \includegraphics[width=0.5\textwidth]{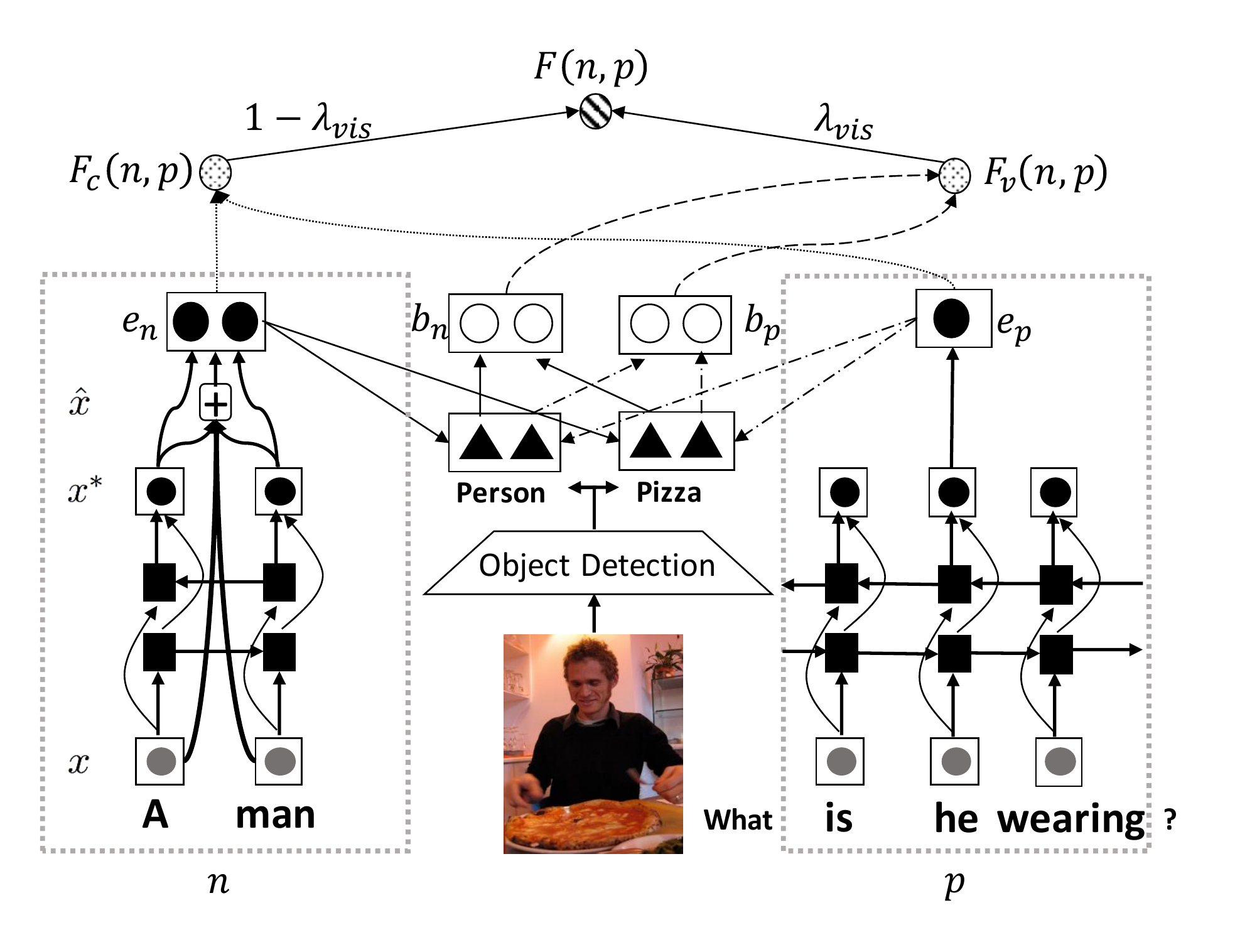}
  \caption{Overall structure of the VisCoref model. Text embedding $e$ is acquired via LSTM and inner-span attention mechanism. Object labels are detected from the image and the possibility $b$ of each text span referring to them is calculated.
  The contextual and visual score $F_c(n,p)$ and $F_v(n,p)$ are calculated upon the features, and the final score $F(n,p)$ is their weighted sum.
  }
  \label{fig:viscoref}
\end{figure}

\section{The Model}\label{sec:model}

The overall framework of the proposed model VisCoref is presented in Figure~\ref{fig:viscoref}.
In VisCoref, we want to leverage both the contextual and visual information to resolve pronouns. Thus we split the scoring function into two parts as follows:
\begin{equation}\label{score}
  F(n, p) = (1 - \lambda_{vis}) \cdot F_c(n, p) + \lambda_{vis} \cdot F_v(n, p),
\end{equation}
where $F_c(\cdot)$ and $F_v(\cdot)$ are the scoring functions based on contextual and visual information respectively. 
$\lambda_{vis}$ is the hyper-parameter to control the importance of visual information in the model.
The details of the two scoring functions are described in the following subsections.

\subsection{Contextual Scoring}
\label{sec:contextual_scoring}
Before computing $F_c$, we first need to encode the contextual information into all the candidates and targeting pronouns through a mention representation module, which is shown as the dotted box in Figure~\ref{fig:viscoref}.

Following \cite{lee-etal-2018-higher}, a standard bidirectional LSTM (BiLSTM) \cite{hochreiter1997long} is used to encode each span with attentions \cite{bahdanau2014neural}.
Assume initial embeddings of words in a span $s_i$ are denoted as $\x_1,...,\x_T$,
and their encoded representation after BiLSTM as $\x^*_1,...,\x^*_T$,
the weighted embeddings of each span $\hat{\x_i}$ are obtained by
\begin{equation}\label{eq:overall_embedding}
  \hat{\x_i} = \sum_{t=1}^{T}a_t \cdot \x_t,
\end{equation}
where $a_t$ is the inner-span attention computed by
\begin{equation}\label{eq:mention_attention}
  a_t = \frac{e^{\alpha_t}}{\sum_{k=1}^{T}e^{\alpha_k}},
\end{equation}
in which $\alpha_t$ is obtained by a standard feed-forward neural network\footnote{We use $NN$ to represent feed-forward neural networks.} $\alpha_t$ = $NN_\alpha(\x^*_t)$.

After that, we concatenate the embeddings of the starting word ($\x^*_{start}$) and the ending word ($\x^*_{end}$) of each span, as well as its weighted embedding ($\hat{\x}_i$) and the length feature ($\phi(i)$)
to form its final representation $\e$:
\begin{equation}\label{eq:mention_embedding}
  \e_i = [\x^*_{start},\x^*_{end},\hat{\x}_i,\phi(i)].
\end{equation}

On top of the extracted mention representation, we then compute the contextual score as follows:
\begin{equation}
    F_c(n,p) = NN_c([\e_p, \e_n, \e_p \odot \e_n]),
\end{equation}
where $[\quad]$ represents the concatenation, $\e_p$ and $\e_n$ are the mention representation of the targeting pronoun and current candidate mention, and $\odot$ indicates the element-wise multiplication.

\subsection{Visual Scoring}
In order to align mentions in the text with objects in the image,
the first step of leveraging the visual information is to recognize the objects from the picture. 
We use a object detection module to identify object labels from each image $I$, such as ``person,'' ``car,'' or ``dog.''
After that, we convert the identified labels into vector representations following the same encoding process in \ref{sec:contextual_scoring}.
For each image, we add a label ``null,'' indicating that the pronoun is referring to none of the detected objects in $I$.
We denote the resulting embeddings as $\e_{c_1}, \e_{c_2}, ..., \e_{c_K}$, in which $c_i$ denotes the detected labels, and $K$ is the total number of unique labels in the corresponding image.

After extracting objects from the image, we would like to know whether the mentions are referring to them.
To achieve this goal, we calculate the possibility of a mention $n_i$ corresponding to each detected object $c_k$:
\begin{equation}
    \beta_{n_i, c_k} = {\rm NN}_\beta\left( {\rm NN}_o(\bm{e}_{n_i}) \odot {\rm NN}_o(\bm{e}_{c_k}) \right).
\end{equation}
Then we take the softmax of $\beta_{n_i, c_k}$ as the final possibility of $n_i$ aligned with the object label $c_k$:
\begin{equation}
    b_{n_i,c_k} = \frac{e^{\beta_{n_i, c_k}}}{\sum_{l={1}}^{K} e^{\beta_{n_i, c_l}}}.
\end{equation}

If $n_i$ corresponds to a certain object in $I$, the score of that label should be large. Otherwise, the possibility of ``null'' should be the largest. Therefore, we use the maximum of possibility scores among all $K$ classes except ``null''
\begin{equation}
    m_i = \max_{k=1,...,K} b_{n_i,c_k}
\end{equation}
as the probability of $n_i$ related to some object in $I$.

Similarly, given two mentions $n_i$ and $n_j$, if they refer to the same detected object $c_k$, then both $b_{n_i,c_k}$ and $b_{n_j,c_k}$ should be large. Thus, we can use the maximum of their product among all K classes except ``null''
\begin{equation}
m_{i,j} = \max_{k=1,...,K} b_{n_i,c_k} \cdot b_{n_j,c_k}
\end{equation}
as the probability of $n_i$ and $n_j$ related to the same object in $I$.

In the end, we define the visual scoring function as follows:
\begin{equation}
    F_v(n, p) = NN_v(\left[m_p, m_n, m_p \cdot m_n, m_{p,n}  \right]).
\end{equation}

\section{The Experiment}\label{sec:experiment}
% 1-1.5 pages
In this section, we introduce the implementation details, experiment setting, and baseline models.

\subsection{Experiment Setting}
As introduced in Section~\ref{sec:post-processing}, we randomly split the data into training, validation, and test sets of 4,000, 500, and 500 dialogues, respectively.
For each dialogue, a mention pool of size 30 is provided for models to detect plausible mentions outside the dialogue. The pool contains both mentions extracted from the corresponding caption and randomly selected negative mention samples from other captions.
All models are evaluated based on the precision (P), recall (R), and F1 score.
Last but not least, we split the test dataset by whether the correct antecedents of the pronoun appear in the dialogue or not.
We denote these two groups as ``Discussed'' and ``Not Discussed.''

\subsection{Implementation Details}
Following previous work \cite{lee-etal-2018-higher}, we use the concatenation of the 300d GloVe embedding~\cite{pennington2014glove} and the ELMo~\cite{peters2018deep} embedding as the initial word representations. 
Out-of-vocabulary words are initialized with zero vectors.
We adopt the ``ssd\_resnet\_50\_fpn\_coco'' model from Tensorflow detection model zoo\footnote{\url{https://github.com/tensorflow/models/tree/master/research/object\_detection}} as the object detection module.
% We adopt the ``ssd\_resnet\_50\_fpn\_coco'' model from Tensorflow detection model zoo\footnote{https://github.com/tensorflow/models/tree/master/ research/object\_detection} as the object detection module.
The size of hidden states in the LSTM module is set to 200, and the size of the projected embedding for computing similarity between text spans and object labels is 512.
The feed-forward networks for contextual scoring and visual scoring have two 150-dimension hidden layers and one 100-dimension hidden layer, respectively.

\begin{table*}[t]
  \centering
    \begin{tabular}{l||ccc|ccc|ccc}
    \toprule
    \multirow{2}[0]{*}{Model}      & \multicolumn{3}{c}{Discussed} & \multicolumn{3}{|c|}{Not Discussed} & \multicolumn{3}{|c}{Overall} \\
     & P     & R     & F1    & P     & R     & F1    & P     & R     & F1 \\
    \midrule
    Deterministic & 61.16 & 25.54 & 36.03 & 17.39 & 8.26  & 11.20 & 56.65 & 24.01 & 33.73 \\
    Statistical & 79.37 & 26.65 & 39.90 & 6.10  & 1.03  & 1.77  & 76.18 & 24.23 & 36.76 \\
    Deep-RL & 72.69 & 26.97 & 39.35 & 32.86 & 14.46 & 20.09 & 68.51 & 25.93 & 37.62 \\
    \midrule
    End-to-end & 89.65 & 63.69 & 74.47 & 67.33 & 64.76 & 66.02 & 86.94 & 63.79 & 73.59 \\
    End-to-end+Visual & 89.78 & 66.31 & 76.28 & 65.17 & 62.38 & 63.75 & 86.91 & 65.95 & 74.99 \\
    \midrule
    Viscoref & 85.95 & 72.45 & \textbf{78.63} & 67.04 & 71.67 & \textbf{69.28} & 83.78 & 72.38 & \textbf{77.66} \\
    \midrule
    Human & 95.03 & 81.82 & 87.93 & 86.18 & 93.57 & 89.73 & 94.02 & 82.91 & 88.12 \\
    \bottomrule
    \end{tabular}%
  \caption{Experimental results on VisPro. Precision (P), recall (R) and F1 scores are presented. The best performed F1 scores are indicated with bold font.}
  \label{tab:main_result}%
\end{table*}%

%\subsection{Training}
For model training, we use cross-entropy as the loss function and Adam \cite{kingma2014adam} as the optimizer.
All the parameters are initialized randomly.
% and we apply dropout rate 0.2 to all hidden layers in the model.
Each mention selects the text span of the highest overall score among all previous text spans in the dialogue or the mention pool as its antecedent, so that all mentions in one dialogue are clustered into several coreference chains. The noun phrases in the same coreference cluster as a pronoun are deemed as the predicted antecedents of that pronoun.
The models are trained with up to 50,000 steps, and the best one is selected based on its performance on the validation set.

\subsection{Baseline Methods}
% Introduce all the evaluated models
Since we are the first to proposed a visual-aware model for pronoun coreference resolution, we compare our results with existing models of general coreference resolution.
\begin{itemize}[leftmargin=*]
\item Deterministic model \cite{raghunathan2010multi} is a rule-based system that aggregates multiple functions for determining whether two mentions are coreferent based on hand-craft features.

\item Statistical model \cite{clark2015entity} learns upon human-designed entity-level features between clusters of mentions to produce accurate coreference chains.

\item Deep-RL model \cite{clark-manning-2016-deep} applies reinforcement learning to mention-ranking models to form coreference clusters.

\item End-to-end model \cite{lee-etal-2018-higher} is the state-of-the-art method of coreference resolution. It predicts coreference clusters via an end-to-end neural network that leverages pretrained word embeddings and contextual information.
\end{itemize}

Last but not least, to demonstrate the effectiveness of the proposed model, we also present a variation of the End-to-end model, which can also use the visual information, as an extra baseline:
\begin{itemize}[leftmargin=*]
    \item End-to-end+Visual first extracts features from images with ResNet-152~\cite{he2016deep}. Then it concatenates the image feature with the contextual feature in the original End-to-end model together to make the final prediction.
\end{itemize}

As the Deterministic, Statistical, and Deep-RL model are included in the Stanford CoreNLP toolkit\footnote{\url{https://stanfordnlp.github.io/CoreNLP/coref.html}}, we use their released model as baselines. For the End-to-end model, we also use their released code\footnote{\url{https://github.com/kentonl/e2e-coref}}.
% As the Deterministic, Statistical, and Deep-RL model are included in the Stanford CoreNLP toolkit\footnote{https://stanfordnlp.github.io/CoreNLP/coref.html}, we use their released model as baselines. For the End-to-end model, we also use their released code\footnote{https://github.com/kentonl/e2e-coref}.

\section{The Result}\label{sec:result}

The experimental results are shown in Table~\ref{tab:main_result}. 
Our proposed model VisCoref outperforms all the baseline models significantly, which indicates that the visual information is crucial for resolving pronouns in dialogues.
Besides that, we also have the following interesting findings:
\begin{enumerate}[leftmargin=*]
    \item For all the conventional models, the ``Not Discussed'' pronouns whose antecedents are absent in dialogues are more challenging than ``Discussed'' pronouns whose antecedents appear in the dialogue context. The reason behind is that if the correct mentions appear in the near context of pronouns, the information about the correct mention can be aggregated to the targeting pronoun through either human-designed rules or deep neural networks (Bi-LSTM). However, when the correct mention is not available in the near context, it is quite challenging for conventional models to understand the dialogue and correctly ground the pronoun to the object both speakers can see, as they do not have the support of visual information.
    % the correct resolution requires a real understanding of the dialogue, which is quite challenging for conventional models without the support of visual information. 
    \item As is shown in the result of the ``End-to-end+Visual'' model, simply concatenating the visual feature to the contextual feature can help resolve ``Discussed'' pronouns but may hurt the performance of the model on ``Not Discussed'' pronouns. Different from them, the proposed Viscoref can improve the resolution of both the ``Discussed'' and ``Not Discussed'' pronouns. There are mainly two reasons behind: (1) The visual information in our model is first converted into textual labels and then transformed into vector representation in the same way as the dialogue context. Thus the vector space of contextual and visual information is perfectly aligned. (2) We introduce a hyper-parameter $\lambda_{vis}$ to balance the influence of different knowledge resources.
    \item Even though our model outperforms all the baseline methods, we still can observe a huge gap between our model and human being. It indicates that current models still cannot fully understand the dialogue even with the support of visual information and further proves the value and necessity of proposing VisPro.
\end{enumerate}

\begin{figure}[t]
    \centering
    \includegraphics[width=\linewidth]{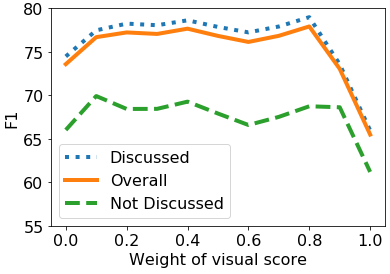}
    \caption{Effect of $\lambda_{vis}$. F1 scores of all categories are reported. 
    % As $\lambda_{vis}$ goes from 0 to 1
    % , the F1 of ``Not Discussed'' pronouns gradually increases, while the score for ``Discussed'' and overall ones goes up to a peak and then drop.
    }
    \label{fig:vis_weight}
\end{figure}

\begin{figure*}[t]
    \centering
    \subfigure[]{\label{fig:example_A}
    \includegraphics[width=0.48\textwidth]{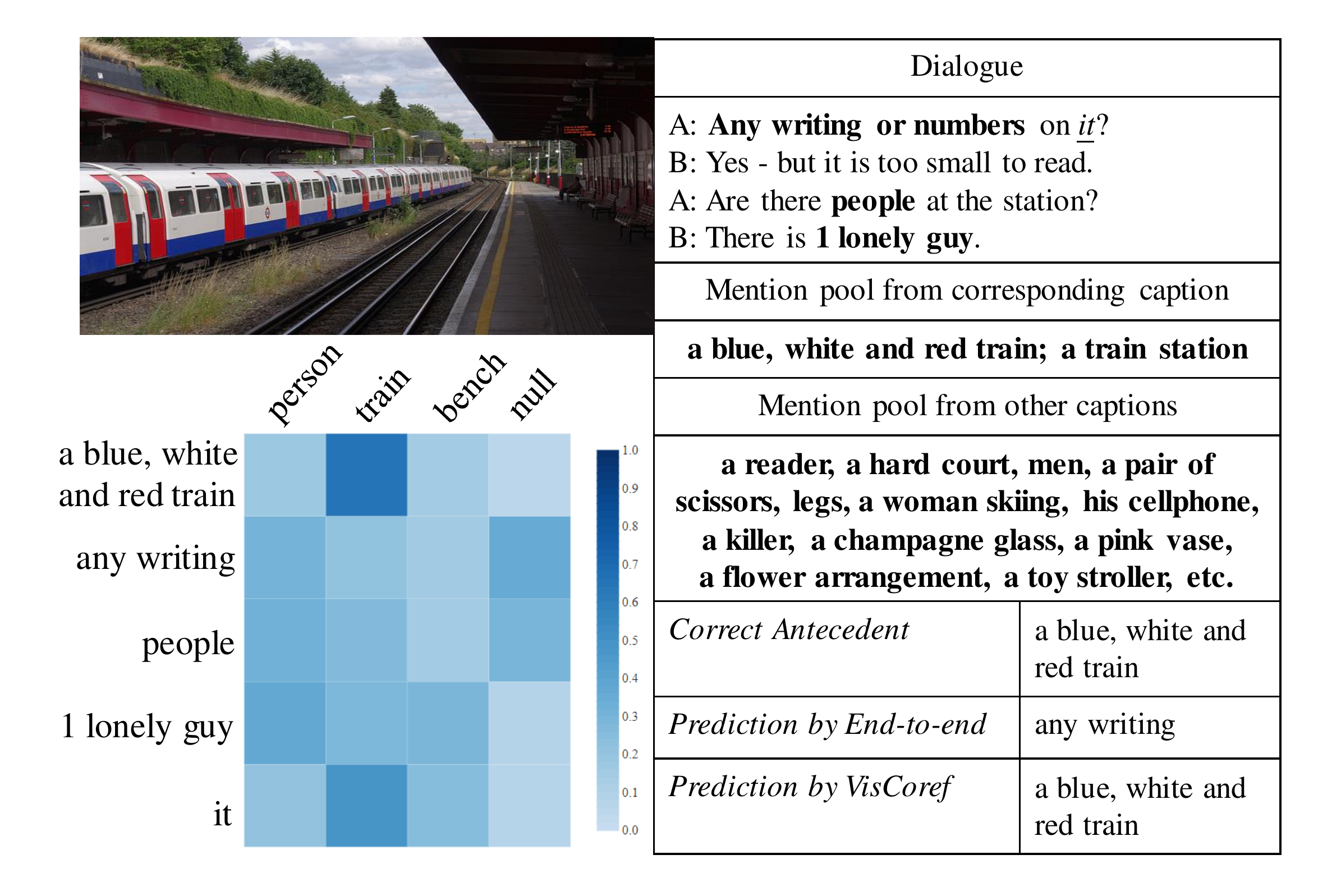}
    }
    \subfigure[]{\label{fig:example_B}
    \includegraphics[width=0.48\textwidth]{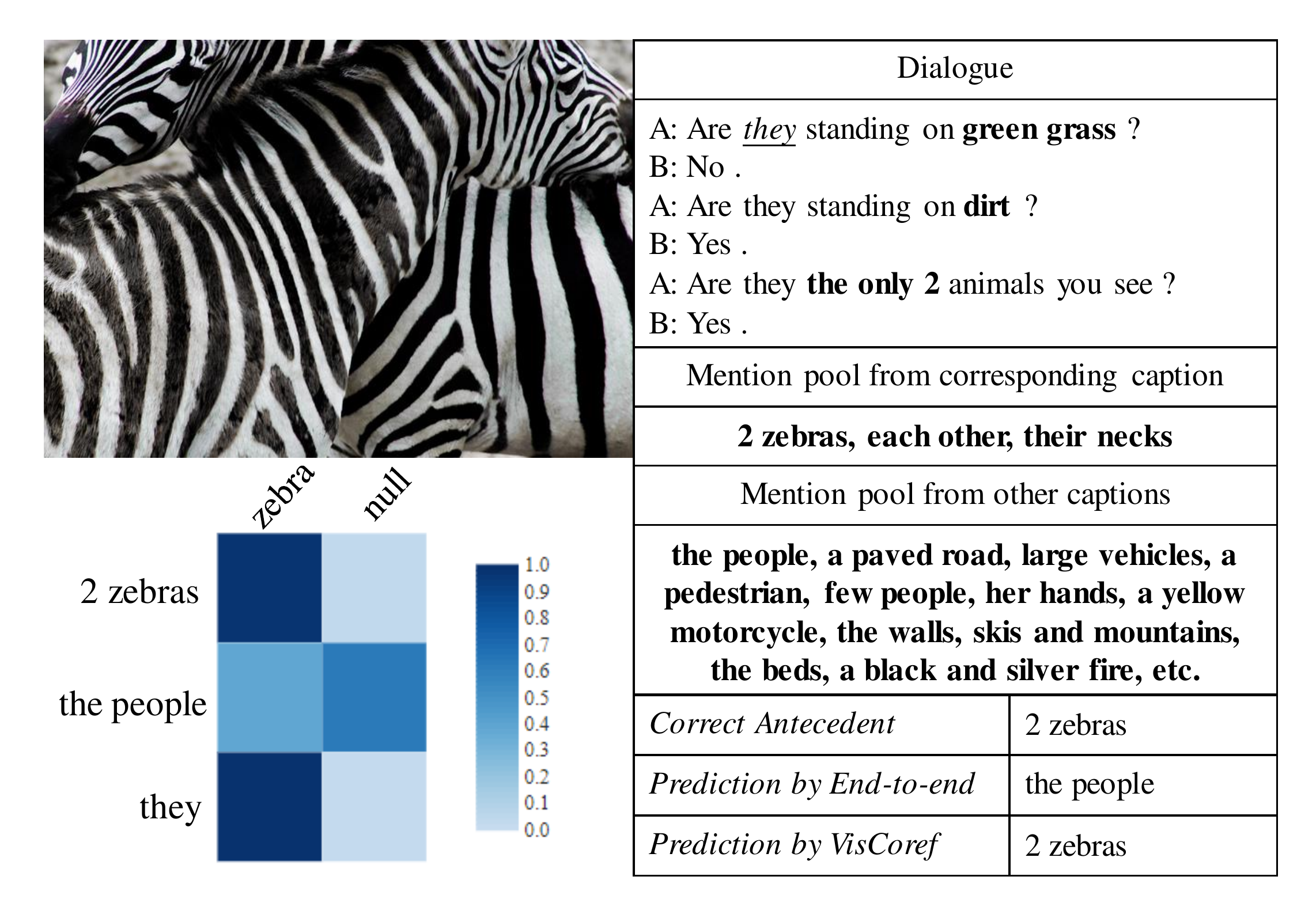}
    }
    \caption{Randomly selected examples from VisPro. The image, dialogue record, prediction result, and heatmap of the mention-object similarity are provided. We indicate the target pronoun with the \underline{\emph{underlined italics}} font and the candidate mentions with \textbf{bold} font. Only relevant parts of dialogues are presented. 
    The row of the heatmap represents mentions in the context, and the column means detected object labels from the image.
    }
    \label{fig:case}
\end{figure*}

\subsection{Hyper-parameter Analysis}

We traverse different weights of visual and contextual information from 0 to 1, and the result is shown in Figure~\ref{fig:vis_weight}. 
Along with the increase of $\lambda_{vis}$, our model puts more weight on the visual information.
As a result, our model can perform better.
However, when our model focuses too much on the visual information (when $\lambda_{vis}$ equals to 0.9 or 1), the model overfits to the visual information and thus performs poorly on the task.
To achieve the balance between the visual and contextual information, we set $\lambda_{vis}$ to be 0.4.

\subsection{Case Study}
To further investigate how visual information can help solve PCR, we randomly select two examples and show the prediction results of VisCoref and End-to-end model in Figure~\ref{fig:case}.

In the first example in Figure~\ref{fig:case}(a), given the pronoun ``it,'' the End-to-end model picks ``any writing'' from the dialogue, while the VisCoref model chooses ``a blue, white and red train'' from the candidate mention sets.
Without looking at the picture, we cannot distinguish between these two candidates. 
However, when the picture is taken into consideration, we observe that there is a train in the image and thus ``a blue, white and red train'' is a more suitable choice, which proves the importance of visual information.
A similar situation happens in Figure \ref{fig:case}(b), where the End-to-end model connects ``they'' to ``the people'' but there is no human being in the image at all.
On the contrary, as VisPro can effectively leverage the visual information and make the decision that ``they'' should refer to ``2 zebras.''

\section{Related Work}\label{sec:related-work}
% 0.75 page
In this section, we introduce the related work about pronoun coreference resolution and visual-aware natural language processing problems.

\subsection{Pronoun Coreference Resolution}

As one core task of natural language understanding, 
pronoun coreference resolution, the task of identifying mentions in text that the targeting pronoun refers to, plays a vital role in many downstream applications in natural language processing, such as machine translation~\cite{guillou-2012-improving}, summarization~\cite{STEINBERGER20071663} and information extraction~\cite{Edens:2003:IBC:860435.860511}.
Traditional studies focus on resolving pronouns in expert-annotated formal textual dataset such as ACE~\cite{nist2003ace} or OntoNotes~\cite{pradhan2012conll}.
However, models that perform well on these datasets might not perform as well in other scenarios such as dialogues due to the informal language and the lack of essential information (e.g., the shared view of two speakers).
In this work, we thus focus on the PCR in dialogues and show that the information contained in the shared view can be crucial for understanding the dialogues and correctly resolving the pronouns.

\subsection{Visual-aware NLP}

As the intersection of computer vision (CV) and natural language processing (NLP), visual-aware NLP research topics have been popular in both communities.
For instance, image captioning~\cite{xu2015show} focuses on generating captions for images, visual question answering (VQA)~\cite{VQA} on answering questions about a image, and visual dialogue~\cite{visdial} on generating a suitable response based on images.
As one vital step of all the aforementioned visual-aware natural language processing tasks~\cite{kottur2018visual}, the visual-aware PCR is still unexplored.
To fill this gap, in this paper, we create VisPro, which is a large-scale visual-aware PCR dataset, and introduce VisCoref to demonstrate how to leverage information hidden in the shared view to resolve pronouns in dialogues better and thus understand the dialogues better.

Another related work is the comprehension of referring expressions \cite{mao2016generation}, which is inferring the object in an image that an expression describes. However, the task is formulated on isolated noun phrases specially designed for discriminative descriptions without putting them into a meaningful context. 
Instead, our task focuses on resolving pronouns in dialogues based on images as the shared view, which enhances the understanding of dialogues based on the comprehension of expressions and images.

\section{Conclusion}\label{sec:conclusion}
In this work, we formally define the task of visual pronoun coreference resolution (PCR) and present VisPro, the first large-scale visual-supported pronoun coreference resolution dataset.
Different from conventional pronoun datasets, VisPro focuses on resolving pronouns in dialogues which discusses a view that both speakers can see. 

Moreover, we also propose VisCoref, the first visual-aware PCR model that aligns contextual information with visual information and jointly uses them to find the correct objects that the targeting pronouns refer to.
Extensive experiments demonstrate the effectiveness of the proposed model.
Further case studies also demonstrate that jointly using visual information and contextual information is an essential path for fully understanding human language, especially dialogues.

\section*{Acknowledgements}

This paper was supported by Beijing Academy of Artificial Intelligence (BAAI), the Early Career Scheme (ECS, No. 26206717) from Research Grants Council in Hong Kong, and the Tencent AI Lab Rhino-Bird Focused Research Program.

\bibliography{emnlp-ijcnlp-2019}
\bibliographystyle{acl_natbib}

\end{document}